\documentclass[lettersize,journal]{IEEEtran}
\usepackage{amsmath,amsfonts}
\usepackage{algorithm}
\usepackage{bm}
\usepackage{algorithmic}
\usepackage{array}
\usepackage[caption=false,font=normalsize,labelfont=sf,textfont=sf]{subfig}
\usepackage{textcomp}
\usepackage{xcolor}
\usepackage{cite}
\usepackage[
    colorlinks=true,
    linkcolor=red,
    citecolor=green,
    urlcolor=blue
]{hyperref}
\usepackage{stfloats}
\usepackage{url}
\usepackage{verbatim}
\usepackage{graphicx}
\usepackage{etoolbox}
\usepackage{booktabs}
\usepackage{multirow}
\usepackage{colortbl}
\definecolor{c1}{RGB}{255,153,153} % best 
\definecolor{c2}{RGB}{255,204,153} % second

\def\BibTeX{{\rm B\kern-.05em{\sc i\kern-.025em b}\kern-.08em
    T\kern-.1667em\lower.7ex\hbox{E}\kern-.125emX}}
\usepackage{balance}

\makeatletter
\def\@IEEEBIOskipN{5\baselineskip}
\expandafter\patchcmd\csname\string\IEEEbiography\endcsname
  {\vskip \@IEEEBIOskipN plus 1fil minus 0\baselineskip}
  {\vskip \@IEEEBIOskipN}
  {}
  {\PackageWarning{main}{Failed to patch IEEEbiography spacing}}
\makeatother

\begin{document}
\title{Relightable 3D Avatar Reconstruction with Semantic-Adaptive Motion-Illumination Responses}
\author{
Jiankuo~Zhao,
Xiangyu~Zhu,~\IEEEmembership{Senior Member,~IEEE},
Jijie~Li,
Baiqin~Wang,
Shukai~Chen,
Zhen~Lei,~\IEEEmembership{Fellow,~IEEE}
\thanks{
Jiankuo Zhao, Xiangyu Zhu, Jijie Li, Baiqin Wang and Zhen Lei are with the State Key Laboratory of Multimodal Artificial Intelligence Systems, Institute of Automation, Chinese Academy of Sciences, Beijing 100190, China, and also with the School of Artificial Intelligence, University of Chinese Academy of Sciences, Beijing 100049, China (e-mail: zhaojiankuo2024@ia.ac.cn; lijijie2024@ia.ac.cn; wangbaiqin2024@ia.ac.cn; xiangyu.zhu@ia.ac.cn; zhen.lei@ia.ac.cn).

Shukai Chen is with ZKTeco Co., Ltd. (e-mail: richard.chen@zkteco.com).
}
}

\markboth{Journal of \LaTeX\ Class Files,~Vol.~18, No.~9, September~2020}%
{How to Use the IEEEtran \LaTeX \ Templates}

\maketitle

\begin{abstract}
Reconstructing expressive and relightable 3D head avatars from monocular videos remains challenging in computer vision, as it requires accurate modeling of both non-rigid facial motion and illumination-dependent appearance. Existing Gaussian avatar methods commonly rely on globally coupled representations, in which Gaussian primitives share a unified motion or illumination response model. Such uniform modeling neglects the distinct motion patterns and material/reflectance properties of different facial semantic regions, thereby limiting fine-grained animation accuracy and reducing relighting plausibility. To address this limitation, we propose SAMIRA, a 3D Gaussian avatar framework for semantic-adaptive motion-illumination response modeling. For motion response modeling, the \textit{Semantic-Adaptive Motion Response} module rasterizes current-to-reference mesh displacements into a topology-consistent UV space and leverages facial semantics to route displacement features through semantic-specific modulators, predicting localized Gaussian geometric residuals beyond coarse mesh binding. For illumination response modeling, the \textit{Semantic-Adaptive Illumination Response} module learns compact diffuse and specular response factors for each facial region, allowing Gaussians in different regions to adapt their illumination responses to novel environment lighting. These response factors are incorporated into deferred physically based shading, providing a lightweight approximation of semantic-dependent illumination effects. Extensive experiments on self-reenactment, cross-reenactment, and relighting demonstrate that SAMIRA improves both fine-grained expression reconstruction and relighting realism over existing methods.
\end{abstract}


\section{Introduction}
\IEEEPARstart{R}{econstructing} expressive and relightable 3D head avatars from monocular videos is a fundamental yet challenging problem in computer vision, with broad applications in virtual and augmented reality, gaming, and film production. A practical avatar system should recover identity-specific geometry and appearance while faithfully modeling how different facial regions deform and react to illumination. We refer to these semantic-specific behaviors as motion-illumination responses: the motion response characterizes semantic-dependent deformation patterns induced by facial expressions, while the illumination response captures semantic-dependent illumination variations under novel lighting. This perspective suggests that animation and relighting cannot be adequately represented by a single global response model, as facial semantic regions differ in both deformation dynamics and interactions with light.

\begin{figure}[t]
	\centering
	\includegraphics[width=1.0\linewidth]{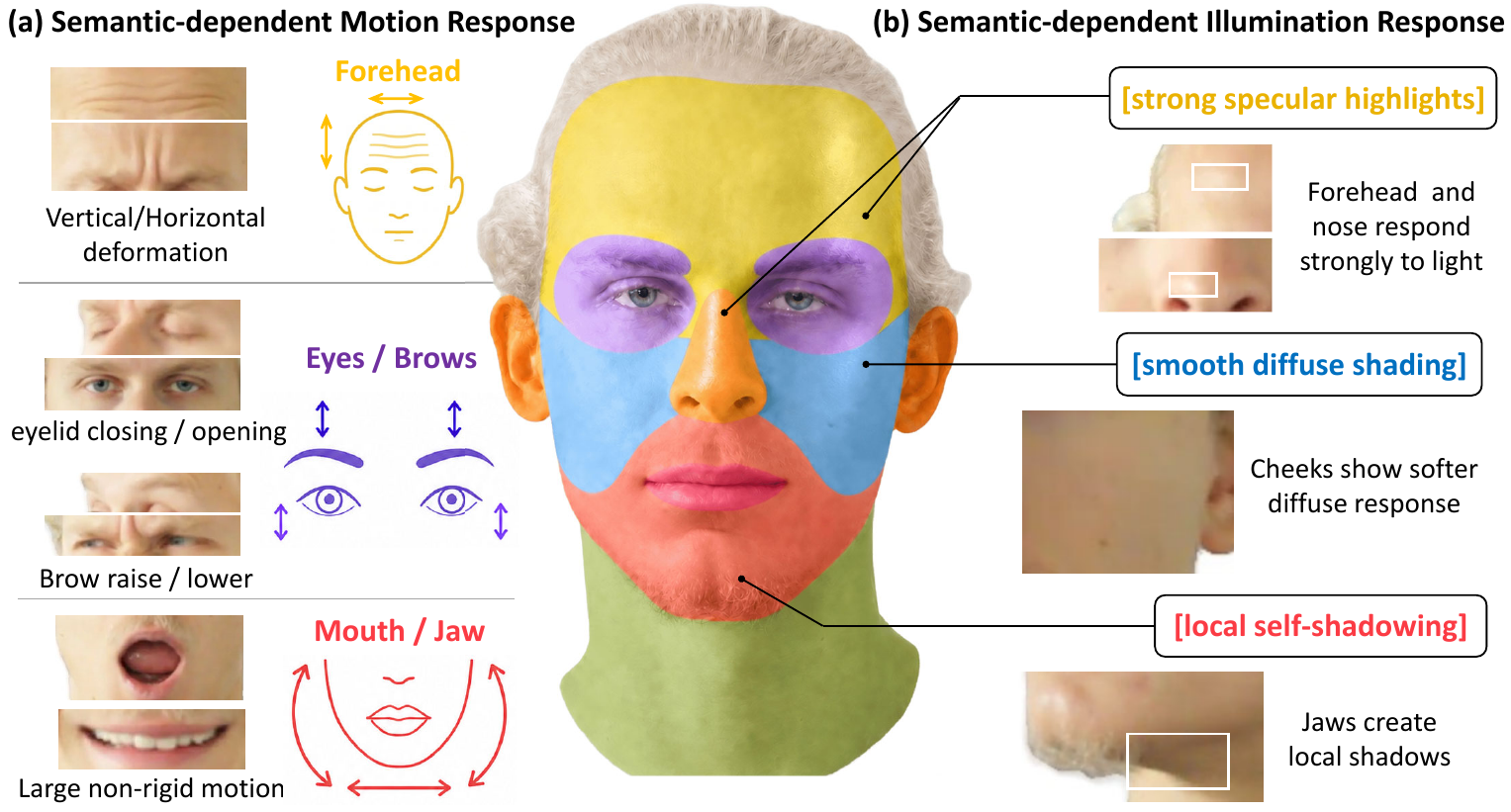}
    \caption{Different facial semantic regions exhibit distinct motion-illumination responses, motivating semantic-adaptive modeling for both local motion response and illumination response.}
	\label{fig:motivation}
\end{figure}

Early head avatar methods often rely on parametric face models, such as 3D Morphable Models (3DMMs)~\cite{blanz2023morphable,paysan20093d}, which enable robust tracking and compact expression control but are constrained by low-dimensional geometry spaces. Neural radiance field methods~\cite{mildenhall2021nerf} improve photorealistic reconstruction through continuous implicit representations; however, they are computationally expensive and often difficult to disentangle geometry, appearance, and illumination. Recently, 3D Gaussian Splatting (3DGS)~\cite{kerbl3Dgaussians} has emerged as an attractive representation by combining explicit primitives, differentiable rasterization, and efficient rendering. These properties make Gaussian avatars well suited for practical monocular reconstruction, but faithfully modeling semantic-specific motion-illumination responses still remains challenging.

For motion response modeling, existing Gaussian avatar methods typically bind Gaussian primitives to a tracked template mesh and animate them via mesh deformation~\cite{qian2023gaussianavatars,SplattingAvatar:CVPR2024,saunders2025gasp}. Although this mesh-bound formulation provides stable global head motion, the residual Gaussian offsets required to recover fine-grained expression details are not spatially random; they exhibit coherent, semantic-dependent motion patterns across facial regions. As illustrated in Fig.~\ref{fig:motivation}(a), human facial expressions arise from localized muscle activations that pull facial tissue in different ways: forehead regions mainly exhibit wrinkle-related vertical or horizontal motion, while mouth and jaw regions involve larger non-rigid deformation, lip separation, and teeth exposure. Although some methods propose neural offsets~\cite{xiang2024flashavatar}, soft binding~\cite{zhao2025stavatar}, and learnable blendshape bases~\cite{li2025rgbavatar,zielonka2025gaussian} to improve deformation capacity, they still rely on globally shared motion modules to process expression signals for the whole face. Such globally coupled motion response modeling can mix regions with incompatible residual directions and scales, making it difficult to consistently recover fine expression details such as wrinkles, eyelid closure, and teeth boundaries.

For illumination response modeling, existing PBR-based avatar methods typically decompose appearance into albedo, normals, material properties, and illumination, thereby enabling rendering under novel lighting conditions. However, different facial regions do not respond to illumination uniformly, since their effective materials, geometric orientations, and local surface variations differ across semantic regions. As illustrated in Fig.~\ref{fig:motivation}(b), regions such as the forehead and nose often produce stronger and more localized specular highlights, whereas the cheeks tend to exhibit smoother diffuse shading and lower-frequency illumination variations. Therefore, a globally shared illumination response model may underfit these region-specific diffuse and specular behaviors, leading to less faithful relighting, especially under directional or high-frequency illumination. This issue becomes more challenging in monocular screen-space supervision, where material properties, normals, and lighting are intrinsically entangled and may compensate for one another during optimization, making region-aware illumination modeling particularly important.

\begin{figure}[t]
	\centering
	\includegraphics[width=1.0\linewidth]{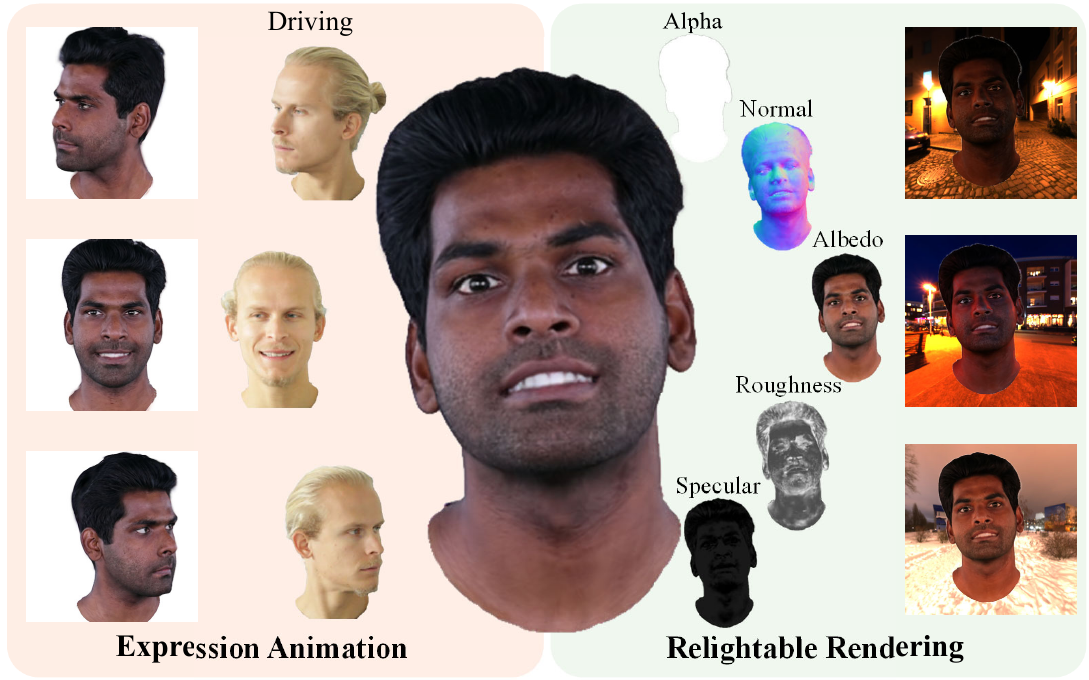}
    \caption{SAMIRA enables accurate expression animation and faithful relighting under novel environment lighting.}
	\label{fig:brief}
\end{figure}

To address these challenges, we propose \textbf{SAMIRA}, a 3D Gaussian avatar framework for semantic-adaptive motion-illumination response modeling, as shown in Fig.~\ref{fig:brief}. For motion response modeling, a \textit{Semantic-Adaptive Motion Response} module rasterizes current-to-reference mesh displacements into a topology-consistent UV space and routes displacement features through region-specific modulators, which predict localized Gaussian geometry residuals beyond coarse mesh binding. For illumination response modeling, a \textit{Semantic-Adaptive Illumination Response} module learns compact diffuse and specular response factors for semantic UV regions and integrates them into deferred PBR Gaussian rendering, enabling different facial regions to adaptively respond to novel illumination. We further introduce a two-stage parameterization strategy to stabilize intrinsic decomposition and the learning of response factors. Experiments on self-reenactment, cross-reenactment, and relighting tasks demonstrate that SAMIRA improves both expression fidelity and relighting realism over existing methods.

Our main contributions are summarized as follows:
\begin{enumerate}[label=\arabic*)]
    \item We present \textbf{SAMIRA}, a Gaussian avatar framework for semantic-adaptive motion-illumination response modeling.
    \item We propose a \textit{Semantic-Adaptive Motion Response} module that complements FLAME-driven mesh binding with UV-aligned displacement features and semantic-specific modulators for fine-scale Gaussian deformation.
    \item We introduce a \textit{Semantic-Adaptive Illumination Response} module that learns compact diffuse and specular response factors for facial regions and integrates them into deferred PBR Gaussian rendering.
    \item We evaluate SAMIRA on expression animation and relighting tasks across multiple datasets, demonstrating improved expression fidelity and relighting realism.
\end{enumerate}
\section{Related Work}

\subsection{3D Head Avatar Reconstruction}
Monocular 3D head avatar reconstruction has evolved from parametric face models~\cite{blanz2023morphable,paysan20093d,li2017learning,yang2020facescape} to implicit neural fields~\cite{mildenhall2021nerf} and, more recently, 3D Gaussian-based representations~\cite{kerbl3Dgaussians}. Parametric models provide robust tracking and compact identity-expression control for monocular reconstruction and animation~\cite{feng2018joint,feng2021learning,retsinas20243d}, but their low-dimensional shape spaces and smooth mesh topology limit fine expression details, mouth interiors, hair, and other non-rigid regions~\cite{zhu2016face,lei2023hierarchical,deng2019accurate}. NeRF-based avatar methods~\cite{gafni2021dynamic,athar2022rignerf,zheng2022avatar,grassal2022neural} improve photorealism with continuous radiance fields, yet usually require costly per-subject optimization and slow volumetric rendering~\cite{hong2022headnerf,INSTA:CVPR2023,kirschstein2023nersemble}.

3D Gaussian Splatting offers an efficient explicit representation with high-quality differentiable rasterization, making it attractive for animatable head avatars. Existing Gaussian avatar methods often bind primitives to a tracked template mesh and use the parametric model as a coarse motion prior~\cite{qian2023gaussianavatars,SplattingAvatar:CVPR2024,tang2025gaf,saunders2025gasp}. While this strategy benefits from stable monocular tracking, purely mesh-driven deformation inherits template limitations and may miss localized motion around the mouth, cheeks, and other high-deformation facial regions. Recent methods therefore predict neural offsets or dynamic Gaussian attributes from pose and expression~\cite{xiang2024flashavatar,chen2024monogaussianavatar,xu2024gaussian,dhamo2024headgas,liao2023hhavatar}, while blendshape-based formulations learn expression-dependent Gaussian bases and weights~\cite{ma20243d,li2025rgbavatar,zielonka2025gaussian}. These designs improve deformation capacity, but their driving signals and motion modules are still mostly shared across semantic facial regions. Our method instead keeps FLAME-guided binding as a robust coarse prior and introduces semantic-adaptive residual modulation for region-specific motion response modeling.

\subsection{Semantic-Aware Facial Modeling}
Semantic priors are widely used to improve local fidelity and controllability in facial modeling. In single-image 3D face reconstruction, facial-part segmentation provides local geometric constraints~\cite{wang20243d}. In neural avatar modeling, NeRFBlendShape~\cite{gao2022reconstructing} builds semantically structured NeRF bases, while local deformation field methods~\cite{chen2023implicit} decompose global motion into locally controllable components. Part-based and semantic-aware implicit representations further demonstrate the value of semantic structural priors for fine-grained facial dynamics~\cite{gu2024parts,jin2024semantic}.

These studies indicate that different facial regions should not necessarily share one deformation model. We extend this idea to mesh-bound Gaussian avatars by using semantic masks as UV-space routing priors. Our design processes displacement cues with semantic-specific modulators and maintains a unified UV residual map for Gaussian animation, preserving stable global mesh binding while assigning different residual capacities to dynamic and relatively stable regions.

\subsection{Avatar Relighting}
Avatar relighting aims to render animatable avatars under novel illumination while preserving identity-specific appearance and details. In monocular settings, it is highly under-constrained because image intensity entangles geometry, albedo, material properties, and lighting. Early high-fidelity systems estimate light transport with dense capture setups or controlled illumination~\cite{guo2019relightables,yang2023towards}, limiting their use in lightweight avatar reconstruction.

A mainstream direction integrates physically based rendering (PBR) into avatar reconstruction. Mesh- and Gaussian-based methods decompose appearance into reusable intrinsic attributes, including albedo, normals, roughness, specular reflectance, and illumination, and perform relighting by changing the illumination in explicit shading models~\cite{bharadwaj2023flare,li2023animatable,HRAvatar,fan2025rgavatar,xu2024relightable}. This improves interpretability and controllability under novel lighting. However, most PBR avatar methods model illumination responses uniformly across facial regions, leaving semantic-dependent diffuse and specular variations underexplored. We therefore augment deferred shading with compact semantic diffuse and specular response factors.

Another line learns relightable appearance or radiance transfer from high-quality multi-view and multi-illumination data, including Relightable Gaussian Codec Avatars and related extensions~\cite{saito2024relightable,li2024uravatar,wang2025relightable}. Although these methods achieve high-quality real-time relighting, they often rely on specialized capture systems, large-scale dynamic data, or strong pretrained avatar models. In contrast, our monocular Gaussian avatar method learns compact diffuse and specular response factors for each semantic facial component, offering a lightweight approximation of semantic-dependent reflectance within deferred PBR shading.

\section{Method}
SAMIRA builds on a mesh-bound Gaussian avatar (Sec.~\ref{sec:prelim}) and introduces two semantic-adaptive response modules. The \textit{Semantic-Adaptive Motion Response} module predicts semantic-specific Gaussian residuals for fine-grained expression control (Sec.~\ref{sec:motion_refinement}), while the \textit{Semantic-Adaptive Illumination Response} module learns compact diffuse and specular response factors for semantic facial regions and integrates them into deferred PBR shading (Sec.~\ref{sec:relight}). Finally, we describe the training objective and optimization strategy (Sec.~\ref{sec:scheme}).

\begin{figure*}[t]
	\centering
	\includegraphics[width=1.0\linewidth]{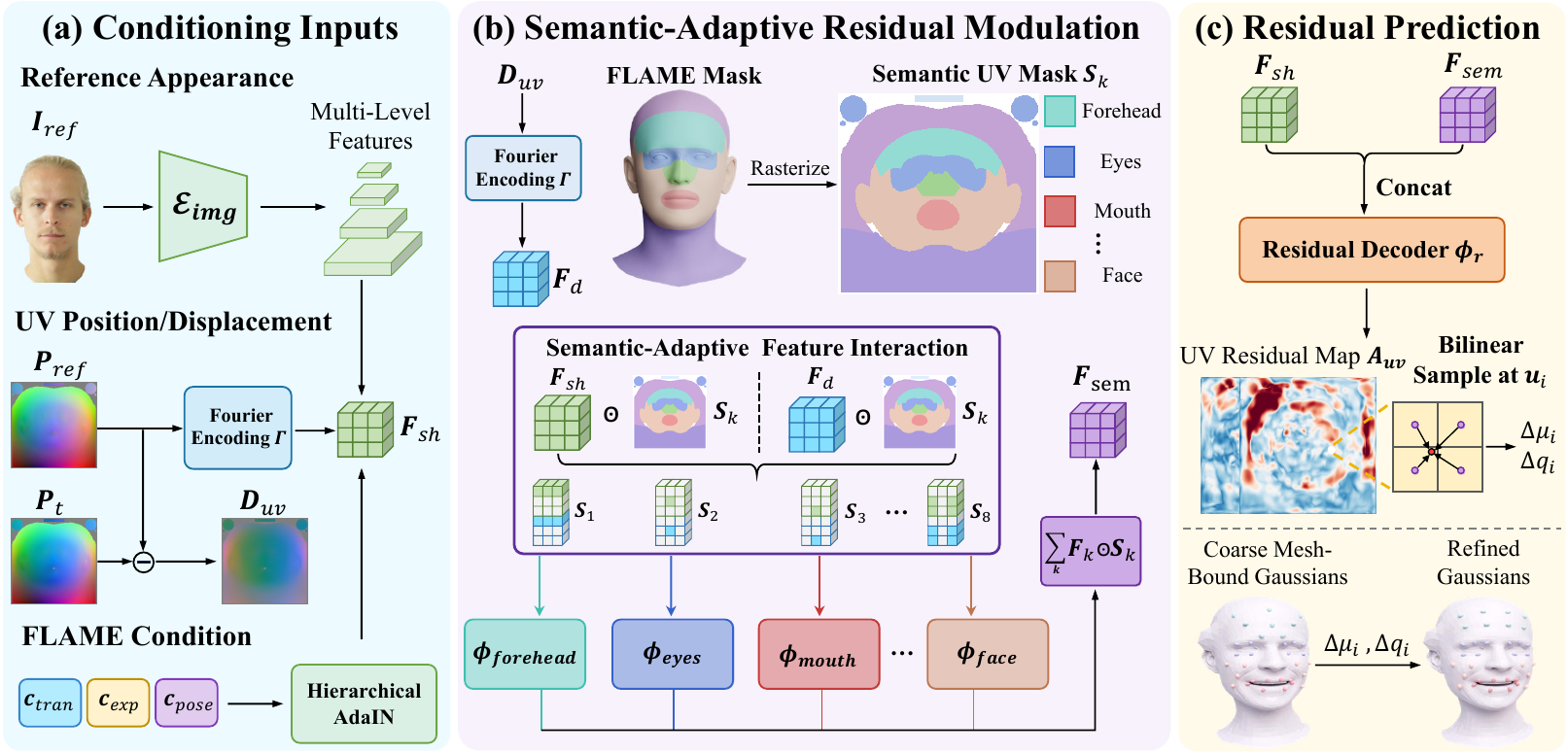}
    \caption{Overview of the proposed Semantic-Adaptive Motion Response module. (a) We first use the reference appearance \(I_{\mathrm{ref}}\), UV position map \(\mathbf{P}_{\mathrm{ref}}\), and FLAME conditions to produce a shared UV-aligned feature \(\mathbf{F}_{\mathrm{sh}}\). (b) UV semantic masks \(\mathbf{S}_k\) route \(\mathbf{F}_{\mathrm{sh}}\) and the displacement feature \(\mathbf{F}_{d}\) through semantic-specific modulators, producing the semantic-adaptive residual feature \(\mathbf{F}_{\mathrm{sem}}\). (c) \(\mathbf{F}_{\mathrm{sem}}\) is concatenated with \(\mathbf{F}_{\mathrm{sh}}\) and decoded into a UV residual map \(\mathbf{A}_{\mathrm{uv}}\), from which each Gaussian samples position and rotation residuals to refine the coarse mesh-bound motion.}
	\label{fig:method}
\end{figure*}

\subsection{Mesh-Bound Gaussian Avatar Preliminaries}
\label{sec:prelim}
Following prior Gaussian-based head avatar reconstruction methods~\cite{qian2023gaussianavatars}, we represent a Gaussian avatar as a set of 3D Gaussians \(\{g_i\}_{i=1}^{N}\) anchored to a tracked FLAME template. Each Gaussian primitive is associated with a mesh triangle in the canonical space and is parameterized by its center \(\bm{\mu}\), anisotropic scale \(\bm{S}\), and rotation matrix \(\bm{R}\). The spatial density of the \(i\)-th Gaussian at position \(\bm{x}\) is defined as
\begin{equation}
    G(\bm{x}) = \exp\!\left(-\frac{1}{2}(\bm{x}-\bm{\mu})^{T}\Sigma^{-1}(\bm{x}-\bm{\mu})\right),
\end{equation}
where the covariance matrix is given by
\begin{equation}
    \Sigma = \bm{R}\,\bm{S}\,\bm{S}^{T}\bm{R}^{T},
\end{equation}
which ensures a positive semi-definite Gaussian distribution. This formulation provides an efficient and differentiable representation for animatable avatars.

To enable animation, each Gaussian is bound to its corresponding mesh triangle and inherits its transformation during mesh deformation. Let \(\bm{r}\) and \(\bm{t}\) denote the rotation and translation of the triangle, respectively, and \(k\) denote a scale factor. The transformed Gaussian parameters are computed as
\begin{equation}
    \bm{\tilde{r}} = \bm{r}\bm{R},\quad
    \bm{\tilde{\mu}} = k\,\bm{r}\bm{\mu} + \bm{t},\quad
    \bm{\tilde{s}} = k\,\bm{s},\quad
    \label{eq:lbs}
\end{equation}
yielding a pose-aligned Gaussian configuration that follows the underlying mesh motion.

This triangle binding strategy provides a robust coarse motion prior from monocular FLAME tracking. However, since all local details are inherited from the parametric mesh deformation, high-motion facial regions such as lips, eyelids, and mouth interiors still require residual corrections beyond triangle-level motion.

\subsection{Semantic-Adaptive Motion Response}
\label{sec:motion_refinement}

\subsubsection{UV-Aligned Conditioning Inputs}
As shown in Fig.~\ref{fig:method}(a), to capture local expression details beyond coarse mesh-bound motion, we construct residual motion conditions in UV space. Compared with screen-space representations, the UV domain preserves the FLAME mesh topology and provides a stable coordinate system for identity-specific residual deformation across facial semantics, poses, and expressions.

We first build two UV-space geometric conditions. The reference UV position map \(\mathbf{P}_{\mathrm{ref}}\) stores, at each valid texel, the canonical mesh-space coordinate of the corresponding point on the FLAME template, thus providing a spatially aligned geometric anchor. For a driven frame, we rasterize the current FLAME mesh \(\mathcal{M}_{t}\) into the same UV layout, yielding the current UV position map \(\mathbf{P}_{t}\), and define the UV displacement map as:
\begin{equation}
    \mathbf{D}_{\mathrm{uv}} = \mathbf{P}_{t} - \mathbf{P}_{\mathrm{ref}}.
    \label{eq:uv_displacement}
\end{equation}

The displacement map \(\mathbf{D}_{\mathrm{uv}}\) describes how the current FLAME mesh deviates from the reference state and serves as the primary frame-dependent residual motion cue. We encode \(\mathbf{D}_{\mathrm{uv}}\) and \(\mathbf{P}_{\mathrm{ref}}\) with Fourier mappings followed by lightweight projections:
\begin{equation}
    \mathbf{F}_{\mathrm{pos}}=\phi_{\mathrm{pos}}\!\left(\Gamma(\mathbf{P}_{\mathrm{ref}})\right), \quad
    \mathbf{F}_{d}=\phi_{d}\!\left(\Gamma(\mathbf{D}_{\mathrm{uv}})\right),
    \label{eq:uv_geom_feat}
\end{equation}
where \(\mathbf{F}_{\mathrm{pos}}\) provides canonical geometric context and \(\mathbf{F}_{d}\) provides frame-dependent residual motion context.

In addition to these UV-space geometric conditions, we extract multi-level appearance features from the reference image \(I_{\mathrm{ref}}\) using a pretrained image encoder \(\mathcal{E}_{\mathrm{img}}\), yielding \(\{\mathbf{F}_{l}^{\mathrm{img}}\}\). Then we inject canonical positional feature \(\mathbf{F}_{\mathrm{pos}}\) into the multi-level image feature pyramid through learnable gates:
\begin{equation}
    \mathbf{F}_{l}^{\mathrm{uv}}
    =
    \mathbf{F}_{l}^{\mathrm{img}}
    +
    \sigma(g_l)\,
    \mathcal{R}_{l}(\mathbf{F}_{\mathrm{pos}}),
    \label{eq:uv_gate}
\end{equation}
where \(g_l\) is a learnable scalar gate for the \(l\)-th feature level, and \(\mathcal{R}_{l}(\cdot)\) resizes the positional feature to the spatial resolution of \(\mathbf{F}_{l}^{\mathrm{img}}\). This gated injection anchors the image features to canonical mesh-space coordinates while allowing the network to adaptively control the strength of geometric conditioning at different scales.

To incorporate the global driving signal into this UV representation, we split the FLAME parameters into translation, expression, and pose-related regions, denoted by \(\mathbf{c}_{\mathrm{tr}}\), \(\mathbf{c}_{\mathrm{exp}}\), and \(\mathbf{c}_{\mathrm{pose}}\), respectively. These components are injected into different feature levels through adaptive normalization:
\begin{equation}
    \bar{\mathbf{F}}_{l}^{\mathrm{uv}}
    =
    \operatorname{AdaIN}_{l}(\mathbf{F}_{l}^{\mathrm{uv}}, \mathbf{c}_{l}),
    \quad
    \mathbf{c}_{l}\in\{\mathbf{c}_{\mathrm{tr}},\mathbf{c}_{\mathrm{exp}},\mathbf{c}_{\mathrm{pose}}\}.
    \label{eq:flame_adain}
\end{equation}

The conditioned pyramid is then fused by a multi-scale module \(\psi\) to produce a shared UV condition feature:
\begin{equation}
    \mathbf{F}_{\mathrm{sh}}
    =
    \psi\!\left(\{\bar{\mathbf{F}}_{l}^{\mathrm{uv}}\}_{l=1}^{L}\right),
    \label{eq:shared_uv_feature}
\end{equation}
which encodes reference appearance, canonical mesh geometry, and global FLAME driving information. Together, the shared UV feature \(\mathbf{F}_{\mathrm{sh}}\) in Eq.~\ref{eq:shared_uv_feature} and the displacement feature \(\mathbf{F}_{d}\) provide the appearance-geometric and residual motion conditions for semantic-partitioned modulation.

\subsubsection{Semantic-Adaptive Residual Modulation}
Although the UV representation provides spatial alignment, directly predicting residual deformation from a globally shared feature may still mix semantic facial regions with distinct motion characteristics. As illustrated in Fig.~\ref{fig:method}(b), to encourage semantic-specific residual modeling, we define \(K\) UV semantic masks \(\{\mathbf{S}_k\}_{k=1}^{K}\), covering regions such as the mouth, eyes, forehead, and other facial parts. These masks partition the UV domain into semantically meaningful areas and serve as fixed spatial priors.

Given the shared UV feature \(\mathbf{F}_{\mathrm{sh}}\) and the displacement feature \(\mathbf{F}_{d}\), we perform semantic-adaptive residual modulation by restricting feature interaction within each semantic mask:
\begin{equation}
    \mathbf{F}_{k}
    =
    \phi_k\!\left(
    [\mathbf{F}_{\mathrm{sh}}\odot\mathbf{S}_k,\;
    \mathbf{F}_{d}\odot\mathbf{S}_k]
    \right),
    \label{eq:region_modulation}
\end{equation}
where \(\odot\) denotes element-wise multiplication, \([\cdot,\cdot]\) denotes channel-wise concatenation, and \(\phi_k\) is a lightweight semantic-specific modulator implemented with grouped convolutions. This design enables displacement cues from different semantic facial regions to be processed by specialized transformations.

The semantic-specific features are then aggregated into a unified semantic-adaptive residual feature:
\begin{equation}
    \mathbf{F}_{\mathrm{sem}}
    =
    \eta\!\left(
    \sum_{k=1}^{K}\mathbf{F}_{k}\odot\mathbf{S}_k
    \right),
    \label{eq:region_feature}
\end{equation}
where \(\eta(\cdot)\) is a lightweight smoothing block that alleviates discontinuities near semantic mask boundaries.

\subsubsection{Residual Prediction}
Given the semantic-adaptive residual feature \(\mathbf{F}_{\mathrm{sem}}\), we predict a UV residual map that refines the coarse mesh-bound Gaussian motion. Specifically, as shown in Fig.~\ref{fig:method}(c), we concatenate \(\mathbf{F}_{\mathrm{sem}}\) with the shared UV feature \(\mathbf{F}_{\mathrm{sh}}\) and decode their joint representation:
\begin{equation}
    \mathbf{A}_{\mathrm{uv}}
    =
    \Phi_r([\mathbf{F}_{\mathrm{sem}}, \mathbf{F}_{\mathrm{sh}}]).
    \label{eq:dynamic_map}
\end{equation}
This formulation keeps the semantic-adaptive residual prediction consistent with the globally conditioned UV representation. The output \(\mathbf{A}_{\mathrm{uv}}\) contains both position and rotation residuals in the UV domain.

For each Gaussian \(g_i\), we sample its geometric residual from \(\mathbf{A}_{\mathrm{uv}}\) according to its UV coordinate \(\mathbf{u}_i\):
\begin{equation}
    [\mathbf{o}_{i}^{\mu},\mathbf{o}_{i}^{r}]
    =
    \mathcal{S}(\mathbf{A}_{\mathrm{uv}},\mathbf{u}_i),
    \label{eq:offset_sample}
\end{equation}
where \(\mathcal{S}(\cdot)\) denotes bilinear sampling. Since Gaussian scale and opacity vary only marginally during facial motion, we optimize them directly as learnable Gaussian parameters rather than explicitly predicting dynamic residuals. To constrain the predicted residual motion, we scale the position residual with factor \(s_{\mu}\) and convert the rotational residual from an axis-angle vector to a quaternion:
\begin{equation}
    \Delta\bm{\mu}_{i}=s_{\mu}\tanh(\mathbf{o}_{i}^{\mu}), \quad
    \Delta\bm{q}_{i}=\operatorname{quat}(\pi\tanh(\mathbf{o}_{i}^{r})).
    \label{eq:offset_process}
\end{equation}

The final Gaussian state is obtained by applying these residuals to the coarse mesh-bound state:
\begin{equation}
    {\bm{\mu}}_{i}^*=\tilde{\bm{\mu}}_{i}+\Delta\bm{\mu}_{i}, \quad
    \bm{q}_{i}^*=\Delta\bm{q}_{i}\otimes\tilde{\bm{q}}_{i}.
    \label{eq:final_motion}
\end{equation}
where \(\tilde{\bm{q}}_{i}\) denotes the quaternion form of the coarse rotation \(\bm{\tilde{r}}_i\) in Eq.~\ref{eq:lbs}. Since the network predicts a UV residual map rather than a fixed list of per-Gaussian residuals, this formulation remains compatible with adaptive density control. The UV coordinate \(\mathbf{u}_i\) of each Gaussian is computed using the geometry-based sampling strategy in Alg.~\ref{alg:geo_uv_sampling}.

\begin{algorithm}[!t]
\caption{Geometry-Based UV Sampling}
\label{alg:geo_uv_sampling}
\begin{algorithmic}[1]
\REQUIRE Pos $\mathbf{P}$, binds $\mathcal{B}$, mesh $(\mathcal{V}, \mathcal{F})$, UV map $(\mathcal{V}_{\mathrm{uv}}, \mathcal{F}_{\mathrm{uv}})$
\ENSURE UV coordinates $\mathcal{U} \in \mathbb{R}^{N \times 2}$
\STATE $\mathcal{T} \gets \mathcal{V}[\mathcal{F}[\mathcal{B}]]$ \COMMENT{Extract bound triangles}
\STATE $\mathbf{B} \gets \mathbf{0}^{N \times 3}$ \COMMENT{Initialize buffer}
\FOR{$i = 1$ \TO $N$}
    \STATE $\mathbf{p} \gets \mathbf{P}[i]$
    \STATE $(\mathbf{a}, \mathbf{b}, \mathbf{c}) \gets \mathcal{T}[i]$
    \STATE Compute dists $d_1 \dots d_6$ for Voronoi regions
    \IF{$\mathbf{p}$ projects to vertex region}
        \STATE $\mathbf{B}_i \gets$ closest vertex weights
    \ELSIF{$\mathbf{p}$ projects to edge region}
        \STATE $\mathbf{B}_i \gets$ 1D edge projection weights
    \ELSE
        \STATE $\mathbf{B}_i \gets$ 2D interior face weights
    \ENDIF
    \STATE $\mathbf{B}_i \gets \max(\mathbf{B}_i, 0)$ \COMMENT{Clamp to valid bounds}
    \STATE $\mathbf{B}_i \gets \mathbf{B}_i / \sum(\mathbf{B}_i)$ \COMMENT{Normalize weights}
\ENDFOR
\STATE $\mathcal{U} \gets \text{BarycentricReweight}(\mathcal{V}_{\mathrm{uv}}, \mathcal{F}_{\mathrm{uv}}, \mathcal{B}, \mathbf{B})$
\end{algorithmic}
\end{algorithm}
\subsection{Semantic-Adaptive Illumination Response}
\label{sec:relight}

\begin{figure*}[th]
	\centering
	\includegraphics[width=0.95\linewidth]{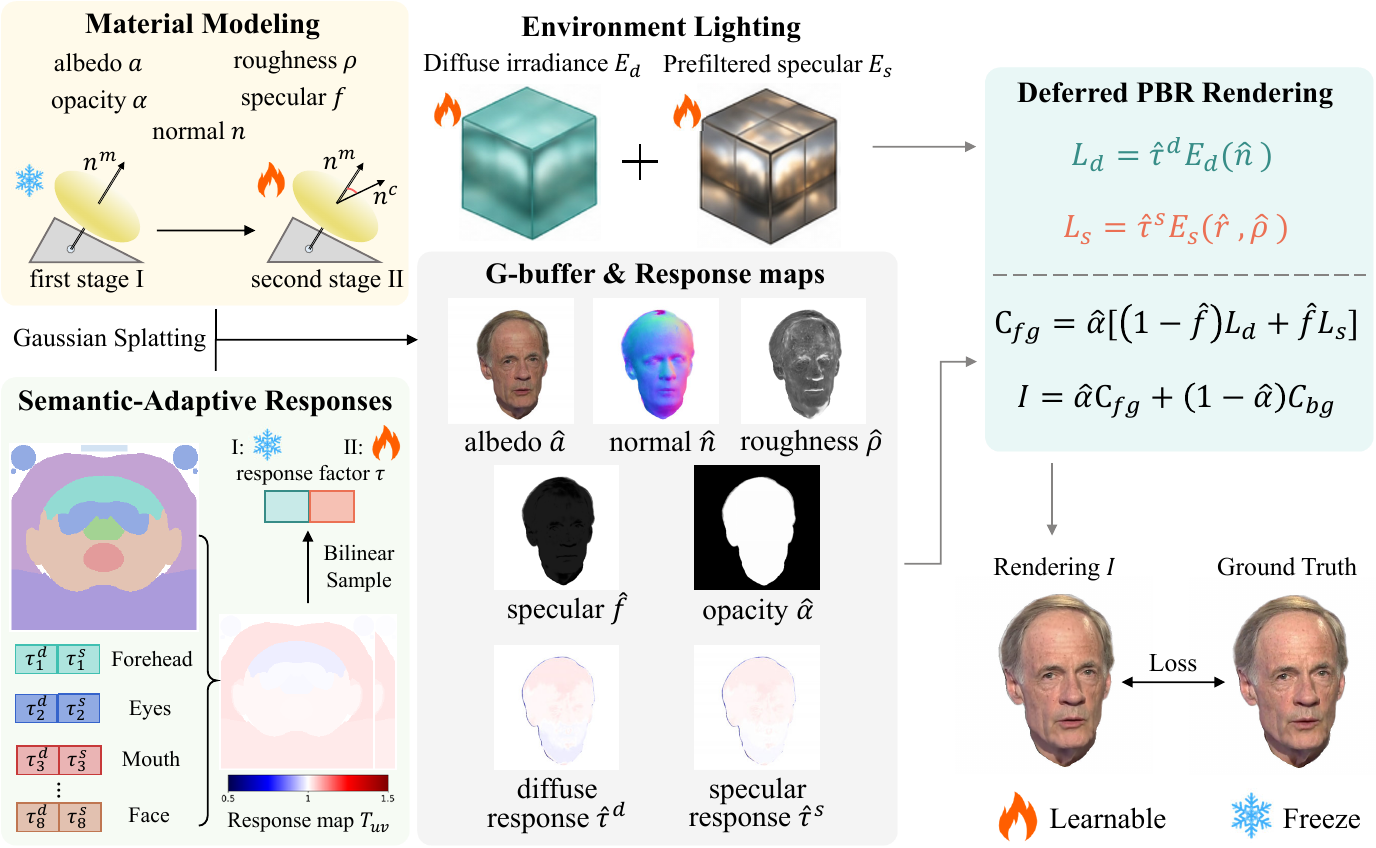}
    \caption{Overview of the proposed Semantic-Adaptive Illumination Response module. We use mesh-aligned Gaussian normals \(\mathbf{n}^{m}\) to stabilize material attribute learning and optimize learnable environment lighting in the first training stage. In the second stage, we further optimize canonical Gaussian normals \(\mathbf{n}^{c}\) for more accurate normal estimation. In addition, each UV semantic area learns compact diffuse and specular response factors \(\bm{\tau}_k=[\tau_k^d,\tau_k^s]\), which are assembled into a response map \(\mathbf{T}^{\mathrm{uv}}\) and sampled by Gaussians. After Gaussian splatting, the rendered G-buffer and response maps modulate diffuse irradiance and prefiltered specular lighting in deferred PBR shading, producing relightable renderings \(\mathbf{I}\).}
	\label{fig:relight_method}
\end{figure*}

\subsubsection{Material and Normal Parameterization}
For relightable rendering, each Gaussian stores albedo \(\mathbf{a}\), roughness \(\rho\), specular reflectance \(\mathbf{f}\), opacity \(\alpha\), and a surface normal, which are rasterized into G-buffers for deferred shading. Since reliable normal supervision is unavailable in monocular videos, we initialize each Gaussian normal from its bound FLAME triangle, denoted as \(\mathbf{n}^{m}\), using the tracked mesh as a stable geometric prior.

To avoid unstable compensation among material, lighting, and normals, we adopt a two-stage parameterization. As shown in the Fig.~\ref{fig:relight_method}, in the first stage, the canonical Gaussian normal is fixed to its mesh-frame initialization \(\mathbf{n}^{m}\), and the semantic response factors are kept neutral while geometry, material attributes, and environment maps are optimized. In the second stage, we activate a learnable canonical normal \(\mathbf{n}^{c}\) together with the response factors, allowing moderate local deviations after the coarse appearance-lighting decomposition becomes stable. For a specific expression, the canonical normal is transformed by the rotation of its bound triangle:
\begin{equation}
    \mathbf{n} =
    \operatorname{norm}(\bm{r}\mathbf{n}^{c}),
    \label{eq:normal_transform}
\end{equation}
where \(\bm{r}\) is defined in Eq.~\ref{eq:lbs}. This keeps the normal consistent with mesh-driven motion while retaining per-Gaussian flexibility in the canonical space. During refinement, normal smoothness and mesh-frame prior losses further regularize the learned normals.

\subsubsection{Semantic-Adaptive Response Factors}
Relightable rendering requires modeling how facial appearance varies under novel illumination. Although illumination is globally defined by the environment map, its visible effects are spatially non-uniform across the face since different semantic facial regions have different effective material and reflectance properties. We therefore introduce a compact semantic-aware approximation that models such systematic variations with facial semantics. As shown in Fig.~\ref{fig:relight_method}, for each UV semantic area \(k\), we learn a two-dimensional illumination response vector:
\begin{equation}
    \bm{\tau}_k=[\tau_k^d,\tau_k^s]
    =
    0.5+\sigma(\bm{\ell}_k),
    \label{eq:region_response}
\end{equation}
where \(\tau_k^d\) and \(\tau_k^s\) modulate the diffuse and specular lighting components, respectively. The logits \(\bm{\ell}_k\) are initialized to zero, yielding a neutral response of \([1,1]\), and are kept frozen during the first stage. In the second stage, we activate \(\bm{\ell}_k\) together with the canonical Gaussian normals and optimize them jointly with the material attributes and environment maps. This schedule lets the response factors learn systematic semantic-dependent illumination behavior after the coarse appearance-lighting decomposition has become stable. We restrict the response range to \((0.5,1.5)\), so that these factors act as mild semantic-level corrections rather than unconstrained lighting terms.

To assign semantic response factors to individual Gaussians, one straightforward strategy is to perform a hard lookup based on the UV semantic mask containing each Gaussian UV coordinate. However, such hard assignment may introduce discontinuities near mask boundaries and requires additional rules to handle overlapping or uncovered masks. We therefore convert the response factors into a dense UV response map through mask-weighted aggregation:
\begin{equation}
    \mathbf{T}^{\mathrm{uv}}(\mathbf{x}) =
    \frac{
    \sum_{k=1}^{K}\mathbf{S}_k(\mathbf{x})\bm{\tau}_k
    }{
    \sum_{k=1}^{K}\mathbf{S}_k(\mathbf{x})+\epsilon
    }.
    \label{eq:uv_response_map}
\end{equation}
where \(\epsilon\) is a small positive constant. For UV pixels not covered by any semantic mask, we assign the neutral response. Each Gaussian then samples \(\mathbf{T}^{\mathrm{uv}}\) at its UV coordinate to obtain \(\bm{\tau}_i=[\tau_i^d,\tau_i^s]\), which is rasterized together with other Gaussian attributes to produce the screen-space response maps \(\hat{\tau}^d\) and \(\hat{\tau}^s\). Compared with hard semantic assignment, this UV-map formulation yields smoother transitions for Gaussians near mask boundaries.

\subsubsection{Deferred PBR Shading and Relighting}
Given the rasterized avatar attributes, we compute the final image with deferred PBR shading. Gaussian splatting produces screen-space G-buffers, including albedo \(\hat{\mathbf{a}}\), normal \(\hat{\mathbf{n}}\), roughness \(\hat{\rho}\), specular reflectance \(\hat{\mathbf{f}}\), opacity \(\hat{\alpha}\), and semantic response maps \(\hat{\tau}^{d}\) and \(\hat{\tau}^{s}\). Since shading is evaluated after rasterization, the learned avatar attributes can be reused under different environment maps at test time.

The semantic response maps are applied to the lighting terms before BRDF evaluation. Specifically, we query a diffuse irradiance map \(E_d(\hat{\mathbf{n}})\) and a prefiltered specular environment map \(E_s(\hat{\mathbf{n}},\hat{\mathbf{v}},\hat{\rho})\), where \(\hat{\mathbf{v}}\) is the view direction. The adjusted lighting is computed as
\begin{equation}
    \mathbf{L}_d
    =
    \hat{\tau}^{d} E_d(\hat{\mathbf{n}}),
    \quad
    \mathbf{L}_s
    =
    \hat{\tau}^{s} E_s(\hat{\mathbf{n}},\hat{\mathbf{v}},\hat{\rho}),
    \label{eq:response_shading}
\end{equation}
where \(\mathbf{L}_d\) and \(\mathbf{L}_s\) denote the semantic-adaptive diffuse and view-dependent specular lighting. This design keeps the PBR renderer standard while allowing each semantic facial component to adapt its diffuse and specular response through compact learned factors.

The foreground color is then evaluated with a standard split-sum BRDF lookup table \((B_x,B_y)\):
\begin{equation}
    \mathbf{C}_{\mathrm{fg}}
    =
    \hat{\mathbf{a}}\odot(1-\hat{\mathbf{f}})\odot\mathbf{L}_d
    +
    \mathbf{L}_s\odot(\hat{\mathbf{f}}B_x+B_y),
    \label{eq:pbr_color}
\end{equation}
where the two terms correspond to diffuse shading and microfacet specular reflection, respectively. The shaded foreground is alpha-composited with the background to obtain the final rendering:
\begin{equation}
    \mathbf{I}
    =
    \hat{\alpha}\mathbf{C}_{\mathrm{fg}}
    +
    (1-\hat{\alpha})\mathbf{C}_{\mathrm{bg}}.
    \label{eq:alpha_compose}
\end{equation}

During training, the diffuse and specular environment maps are learnable cubemaps, with the specular map filtered into roughness-dependent mip levels. At test time, relighting is performed by replacing these cubemaps with external environment maps while keeping the learned Gaussian material attributes, normals, and semantic-adaptive illumination responses fixed.

\begin{table*}[t]
    \centering
    \caption{Quantitative self-reenactment comparison on INSTA, HDTF, and NerFace datasets. We report PSNR, SSIM, LPIPS, and \(L_1^*\) (\(\times10^2\)); the \colorbox{c1}{best} and \colorbox{c2}{second-best} results are highlighted.}
    \resizebox{\textwidth}{!}
    {
        \fontsize{9pt}{11pt}\selectfont
        \begin{tabular}{ccccccccccccc}
            \toprule
            \multirow{2}{*}{Method} & \multicolumn{4}{c}{INSTA Dataset} & \multicolumn{4}{c}{HDTF Dataset} & \multicolumn{4}{c}{NerFace Dataset} \\
            \cmidrule(lr){2-5} \cmidrule(lr){6-9} \cmidrule(lr){10-13}
            & PSNR\(\uparrow\) & SSIM\(\uparrow\) & LPIPS\(\downarrow\) & \(L_1^*\)\(\downarrow\) & PSNR\(\uparrow\) & SSIM\(\uparrow\) & LPIPS\(\downarrow\) & \(L_1^*\)\(\downarrow\) & PSNR\(\uparrow\) & SSIM\(\uparrow\) & LPIPS\(\downarrow\) &\(L_1^*\)\(\downarrow\)\\
            \midrule
            FLARE \cite{bharadwaj2023flare} & 25.44 & 0.9115 & 0.0703 & 1.417 & 23.46 & 0.8633 & 0.1045 & 2.249 & 24.76 & 0.9104 & 0.0799 & 1.625 \\
            SplattingAvatar \cite{SplattingAvatar:CVPR2024} & 27.48 & 0.9329 & 0.1046 & 1.326 & 24.05 & 0.8674 & 0.1967 & 2.690 & 26.14 & 0.9238 & 0.1145 & 1.563 \\
            GaussianAvatars \cite{qian2023gaussianavatars} & 26.99 & 0.9378 & 0.0851 & 1.305 & 23.46 & 0.8742 & 0.1685 & 2.573 & 25.74 & 0.9279 & 0.0965 & 1.527 \\
            FlashAvatar \cite{xiang2024flashavatar} & 27.87 & 0.9354 & 0.0563 & 1.194 & 25.49 & 0.8880 & 0.0827 & 1.938 & 26.96 & 0.9331 & 0.0555 & 1.347 \\
            FateAvatar \cite{zhang2025fate} & 28.33 & 0.9446 & 0.0508 & 1.061 & 25.61 & 0.8963 & 0.0903 & 1.851 & 27.12 & 0.9364 & 0.0558 & 1.266 \\
            HRAvatar \cite{HRAvatar} & \cellcolor{c2}28.62 & \cellcolor{c2}0.9512 & \cellcolor{c2}0.0403 & \cellcolor{c2}0.948 & \cellcolor{c2}26.72 & \cellcolor{c2}0.9213 & \cellcolor{c2}0.0589 & \cellcolor{c1}1.446 & \cellcolor{c2}27.89 & \cellcolor{c2}0.9485 & \cellcolor{c2}0.0407 & \cellcolor{c2}1.096 \\
            \midrule
            Ours & \cellcolor{c1}30.50 & \cellcolor{c1}0.9579 & \cellcolor{c1}0.0307 & \cellcolor{c1}0.806 & \cellcolor{c1}27.47 & \cellcolor{c1}0.9224 & \cellcolor{c1}0.0525 & \cellcolor{c2}1.451 & \cellcolor{c1}30.13 & \cellcolor{c1}0.9570 & \cellcolor{c1}0.0291 & \cellcolor{c1}0.907 \\
            \bottomrule
        \end{tabular}
    }
    \label{tab:recon}
\end{table*}

\subsection{Training Objective and Optimization}
\label{sec:scheme}

\subsubsection{Training Objective}
We optimize SAMIRA with an RGB reconstruction loss and two intrinsic regularizers. The RGB term combines \(\mathcal{L}_1\), D-SSIM, and a perceptual loss that is activated in the latter half of training:
\begin{equation}
	\mathcal{L}_{\mathrm{rgb}}
	=
	(1-\lambda_1)\mathcal{L}_1+\lambda_1\mathcal{L}_{\mathrm{DSSIM}}
	+\gamma\lambda_2\mathcal{L}_{\mathrm{LPIPS}},
	\label{eq:rgb}
\end{equation}
where \(\lambda_1=0.2\), \(\lambda_2=0.05\), and \(\gamma=1\) only when the perceptual loss is enabled and \(0\) otherwise.

For normal refinement, we combine local angular smoothness on rendered normals with a mesh-frame prior on canonical Gaussian normals:
\begin{equation}
	\begin{gathered}
		\mathcal{L}_{n}^{sm}
		=
		\frac{1}{|\mathcal{P}|}
		\sum_{(p,q)\in\mathcal{P}}
		\left(1-\hat{\mathbf{n}}_{p}^{T}\hat{\mathbf{n}}_{q}\right),\\
		\mathcal{L}_{n}^{prior}
		=
		\frac{1}{N}
		\sum_{i=1}^{N}
		\left(1-(\mathbf{n}^{c}_{i})^{T}\bar{\mathbf{n}}^{c}_{i}\right),\\
		\mathcal{L}_{\mathrm{normal}}
		=
		\lambda_3\mathcal{L}_{n}^{sm}
		+
		\lambda_4\mathcal{L}_{n}^{prior},
	\end{gathered}
	\label{eq:normal_loss}
\end{equation}
where \(\mathcal{P}\) denotes neighboring pixel pairs, \(N\) is the number of Gaussians, and \(\bar{\mathbf{n}}^{c}_{i}\) is the mesh-frame normal initialization of the \(i\)-th Gaussian. We set \(\lambda_3=0.15\) and \(\lambda_4=0.1\).

For albedo regularization, we use a pseudo albedo map \(\mathbf{A}^{*}\) extracted by~\cite{chen2024intrinsicanything} to reduce baked-in illumination:
\begin{equation}
	\mathcal{L}_{\mathrm{alb}}
	=
	\lambda_5 \|\hat{\mathbf{a}}-\mathbf{A}^{*}\|_1 ,
	\label{eq:albedo_loss}
\end{equation}
where \(\lambda_5=0.1\).

Overall, we define the training objective by combining the RGB reconstruction loss and the intrinsic regularization terms:
\begin{equation}
	\mathcal{L}_{\mathrm{total}}
	=
	\mathcal{L}_{\mathrm{rgb}}+\mathcal{L}_{\mathrm{normal}}+\mathcal{L}_{\mathrm{alb}}.
	\label{eq:total_loss}
\end{equation}

\subsubsection{Optimization}
We employ Adam~\cite{kingma2014adam} for all learnable parameters. The Semantic-Adaptive Motion Response network is trained with a learning rate decayed from \(1\times10^{-4}\) to \(1\times10^{-5}\), while the learnable environment maps are trained from \(2\times10^{-4}\) to \(2\times10^{-5}\) using the same warm-up cosine schedule. We further finetune the per-frame FLAME expression, translation, and pose parameters during training. During the first \(30\%\) of training, canonical normals and semantic-adaptive illumination response factors are frozen to their neutral initialization; they are then activated and optimized jointly for the remaining iterations. We follow the adaptive density control of 3DGS for densification and pruning, and update Gaussian UV sampling coordinates after densification to keep the UV residual and response maps aligned with the current Gaussian set.

\section{Experiments}

\begin{figure*}[ht]
	\centering
	\includegraphics[width=1.0\linewidth]{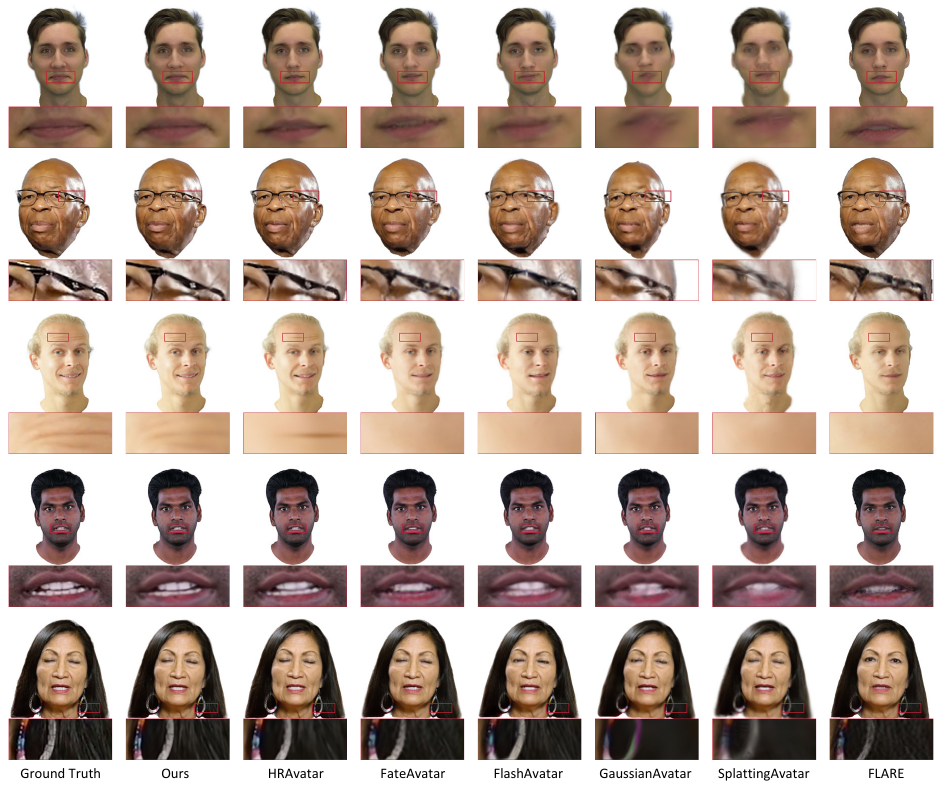}
    \caption{Qualitative self-reenactment comparison. Our method better preserves identity details while producing more faithful local expression dynamics, such as facial wrinkles, and clearer fine structures, including earrings, eyeglasses, and teeth.}
	\label{fig:recon}
\end{figure*}

\subsection{Experimental Setup}
\label{sec:settings}

\subsubsection{Implementation Details}
For each identity, SAMIRA is trained for 10 epochs with the two-stage parameterization described in Sec.~\ref{sec:relight}. Input frames are cropped and resized to \(512\times512\). We use foreground alpha mattes obtained with Robust Video Matting~\cite{lin2022robust} and composite the subject onto a white background for both training and evaluation. FLAME parameters and camera calibration are estimated using VHAP~\cite{qian2024vhap}, which provides the coarse mesh motion. UV position maps, displacement maps, and semantic masks are constructed at a resolution of \(256\times256\). For relightable appearance learning, the learnable environment maps and Gaussian material attributes are jointly optimized with the avatar representation. All experiments are conducted on a single NVIDIA RTX 3090 GPU.

\subsubsection{Datasets}
We evaluate the proposed method on three publicly available monocular video datasets: INSTA~\cite{INSTA:CVPR2023}, HDTF~\cite{zhang2021flow}, and NerFace~\cite{gafni2021dynamic}, covering 25 identities in total. For each sequence, we follow the same preprocessing pipeline and reserve the last 450 frames for testing, while the remaining frames are used for subject-specific optimization.

\subsubsection{Baselines and Training Protocol}
We compare our approach with six state-of-the-art avatar reconstruction methods: GaussianAvatars~\cite{qian2023gaussianavatars}, FateAvatar~\cite{zhang2025fate}, HRAvatar~\cite{HRAvatar}, FlashAvatar~\cite{xiang2024flashavatar}, SplattingAvatar~\cite{SplattingAvatar:CVPR2024}, and FLARE~\cite{bharadwaj2023flare}. For fairness, all methods are evaluated on the same train/test splits. Since different baselines have different convergence behaviors and training schedules, we follow their official configurations rather than forcing an identical epoch number.

\subsection{Self-Reenactment Evaluation}
We first evaluate self-reenactment quality using PSNR, SSIM, LPIPS~\cite{zhang2018unreasonable}, and \(L_1^*\) distance. As reported in Tab.~\ref{tab:recon}, SAMIRA achieves the best overall performance on the three datasets, with especially clear gains in LPIPS and L1* on INSTA and NerFace. These improvements indicate that the reconstructed avatars better preserve perceptually important expression details, rather than merely increasing pixel-wise similarity. The gains are also consistent across identities and motion patterns, supporting the claim that semantic-adaptive motion response modeling improves expression control under monocular reconstruction.

Fig.~\ref{fig:recon} further shows qualitative comparisons. Compared with mesh-bound or globally conditioned baselines, SAMIRA produces sharper facial structures and more faithful local expression details, including forehead wrinkles and subtle changes around the mouth corners. These areas are difficult to model with a single globally coupled deformation since their motion amplitudes and local appearance changes differ from relatively stable facial regions. By treating FLAME motion as a coarse prior and predicting UV-aligned Gaussian residuals, our method better aligns lips, eyelids, and neighboring skin areas under varying expressions. The improved local alignment also helps preserve identity-related details and accessories such as glasses and earrings.

\begin{figure*}[thb]
	\centering
	\includegraphics[width=1.0\linewidth]{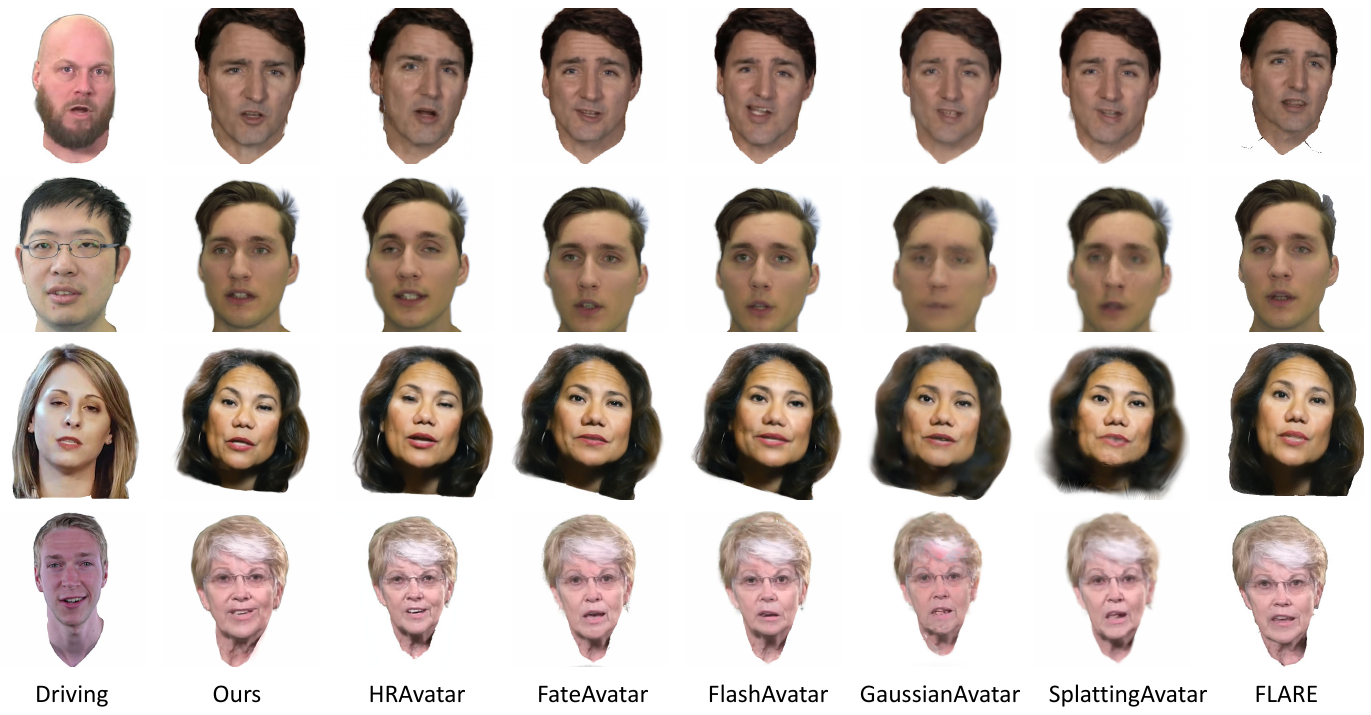}
	\caption{Qualitative cross-reenactment results. Our method more faithfully reproduces the driving identity's expressions, including eye closure and smiling.}
	\label{fig:cross}
\end{figure*}

\subsection{Cross-Reenactment Evaluation}
We further evaluate cross-identity reenactment by driving a reconstructed source avatar with expressions from another identity. As shown in Fig.~\ref{fig:cross}, our method accurately transfers expressions such as smiling while maintaining the source identity appearance. The Semantic-Adaptive Motion Response module provides local flexibility beyond the coarse FLAME binding, allowing highly dynamic facial regions to follow the target expression more closely while keeping relatively stable regions less affected. This is especially clear around the lips and eyelids, where different identities may exhibit different semantic-specific motion patterns under similar FLAME parameters.

\subsection{Relighting Evaluation}
\subsubsection{Relighting Comparisons}
\begin{figure*}[h]
	\centering
	\includegraphics[width=1.0\linewidth]{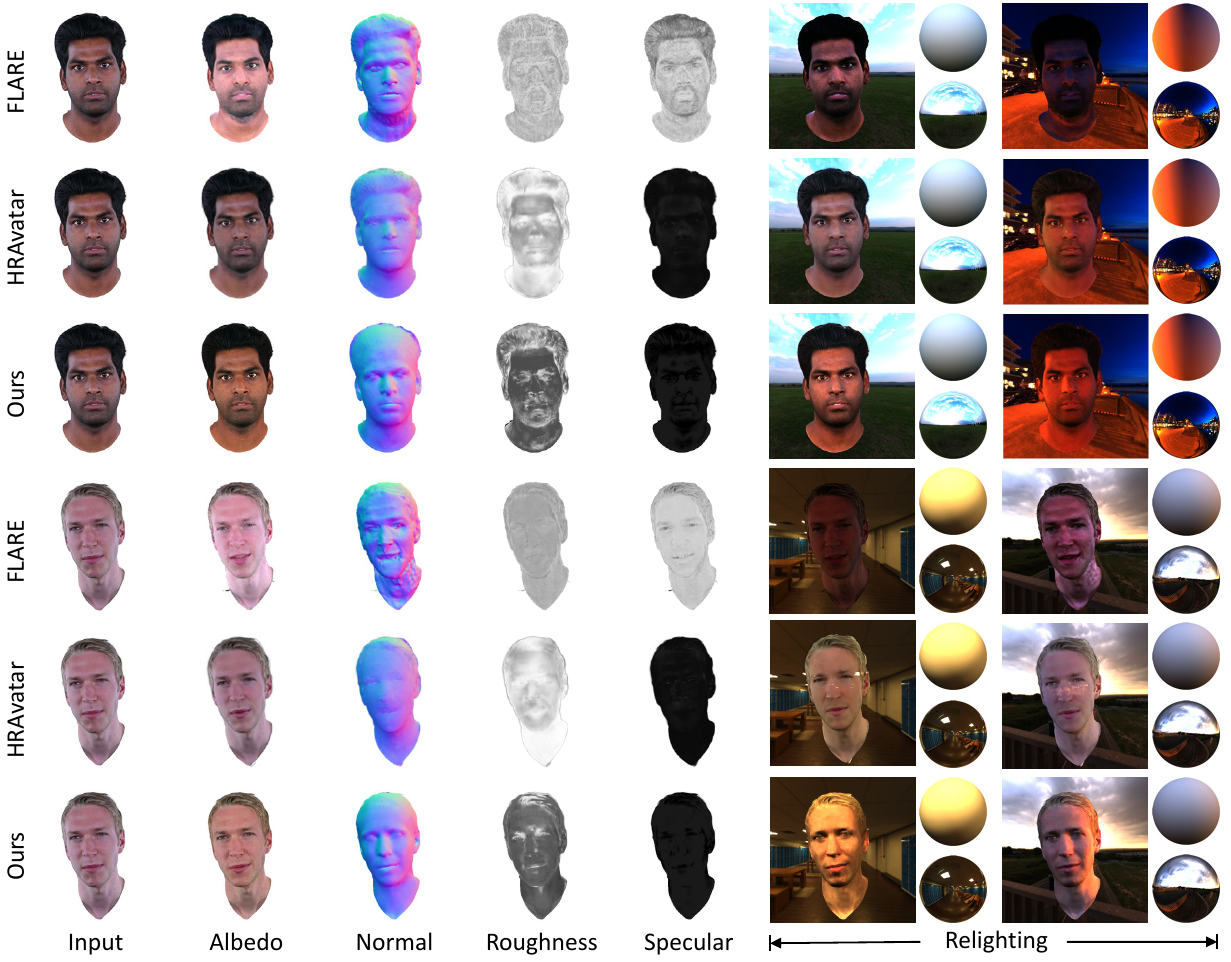}
	\caption{We visualize the albedo, normal, roughness, specular, and relit renderings produced by three relightable avatar methods: FLARE, HRAvatar, and SAMIRA. Ours produces more accurate normal estimation and achieves more faithful relighting results under novel illumination.}
	\label{fig:relight}
\end{figure*}

We evaluate relightable rendering by comparing SAMIRA with two relightable avatar reconstruction methods, FLARE and HRAvatar, under novel environment maps. As shown in Fig.~\ref{fig:relight}, we visualize the intrinsic attributes and relit results for each method. The light probes and shaded spheres indicate the target illumination used for relighting. SAMIRA produces more plausible illumination changes across different lighting conditions. Under outdoor and night-time environment maps, our renderings show clearer directional shading, stronger but controlled highlights, and more natural shadow variations around facial structures such as the nose, cheeks, and eye sockets. In contrast, FLARE often yields less consistent facial shading, while HRAvatar tends to produce flatter or overly dark relit appearances in some areas. The improved realism is closely related to the semantic-adaptive response factors, which enables SAMIRA to better reflect semantic-dependent material responses while preserving identity details under novel illumination. Besides, our normal maps preserve more coherent facial geometry and local surface orientation, which helps the deferred PBR renderer place highlights and shadows in physically more plausible locations.

\begin{figure}[t]
	\centering
	\includegraphics[width=1.0\linewidth]{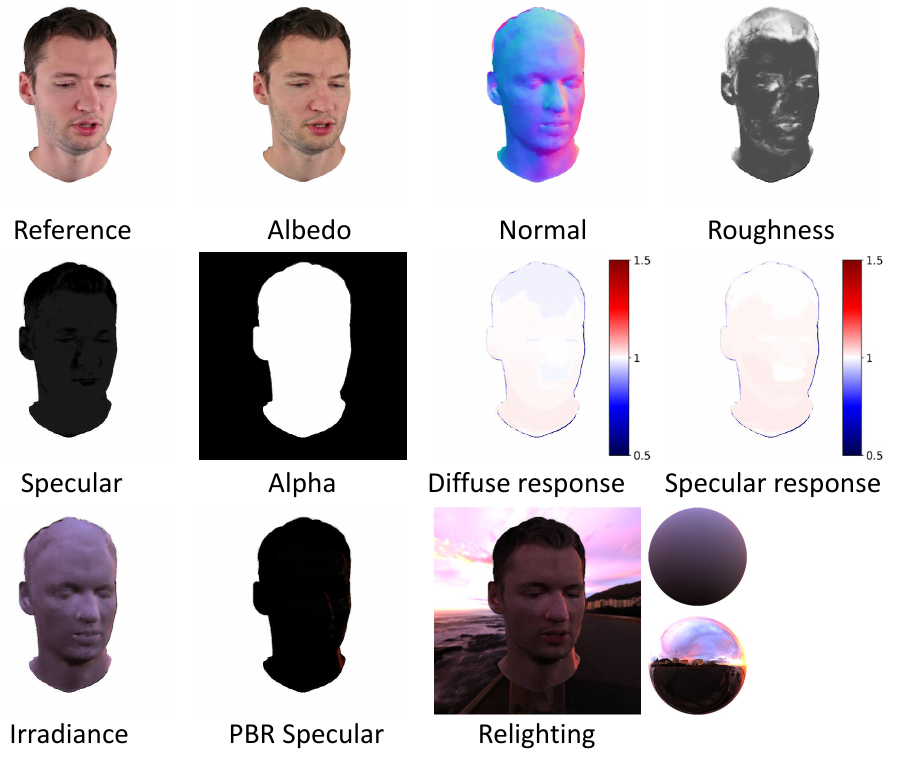}
	\caption{Visualization of SAMIRA's decomposed attributes, semantic response maps, deferred lighting components, and relighting result.}
	\label{fig:material}
\end{figure}

\subsubsection{Material Decomposition}
Fig.~\ref{fig:material} visualizes the attributes and lighting components produced by SAMIRA, including albedo, normal, roughness, specular reflectance, alpha, and the learned semantic response maps. The response maps remain close to the neutral value while exhibiting spatial variations across facial semantics, suggesting that they serve as compact semantic-level illumination corrections rather than unconstrained lighting textures. Specifically, the diffuse response primarily adjusts broad, low-frequency illumination, whereas the specular response modulates localized view-dependent highlights. The bottom row shows the deferred shading components under a novel environment map. The irradiance map captures smooth diffuse lighting over the face, while the PBR specular component produces stronger highlights on the right side, consistent with the input illumination. These components are reflected in the final relit rendering, demonstrating that the learned attributes and semantic response factors provide interpretable inputs for relightable PBR rendering.

\subsection{Ablation Study}

\subsubsection{Semantic-Adaptive Motion Response}
\begin{figure}[t]
	\centering
	\includegraphics[width=1.0\linewidth]{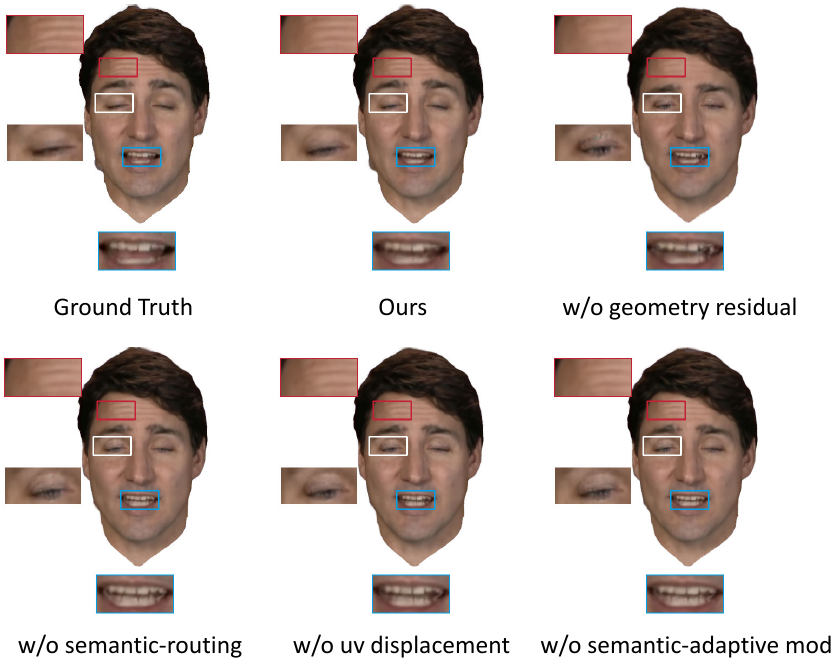}
	\caption{Qualitative ablation results. The highlighted areas show that ours better preserves local details under expression changes.}
	\label{fig:ablation}
\end{figure}

\begin{figure}[t]
	\centering
	\includegraphics[width=1.0\linewidth]{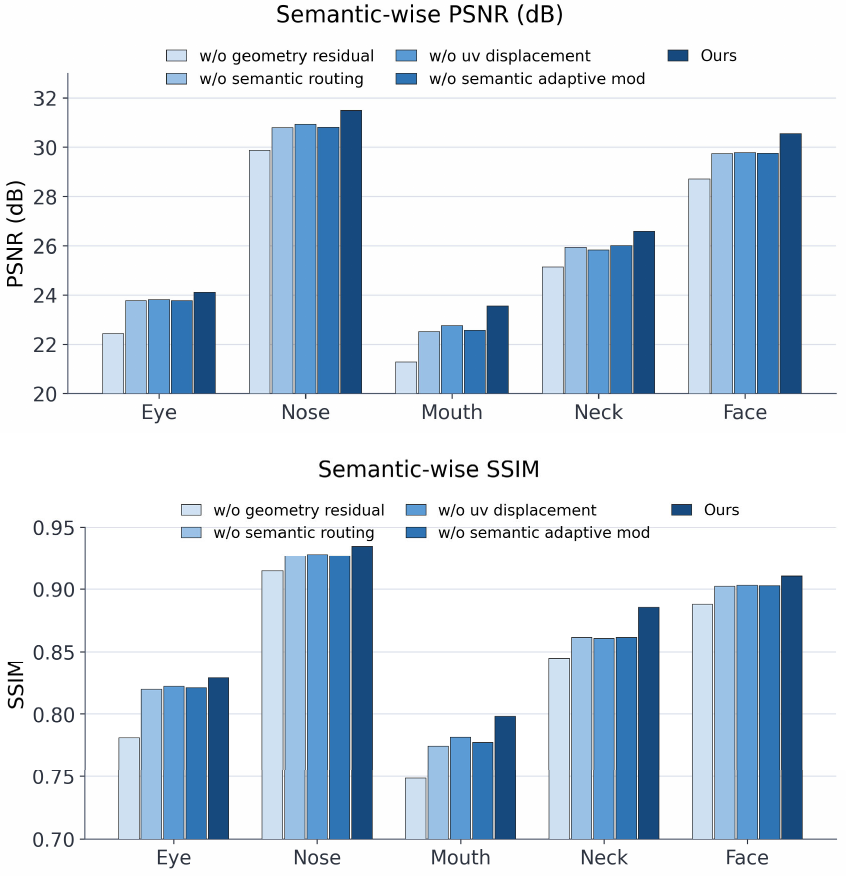}
	\caption{Semantic-wise PSNR and SSIM for motion-response ablations.}
	\label{fig:ablation_region}
\end{figure}

We first evaluate the design choices of the Semantic-Adaptive Motion Response module. Tab.~\ref{tab:ablation} summarizes the overall reconstruction metrics, Fig.~\ref{fig:ablation} presents qualitative comparisons of local details, and Fig.~\ref{fig:ablation_region} reports per-semantic-class PSNR and SSIM for a more comprehensive analysis.

\begin{table}[t]
	\centering
	\caption{Quantitative ablation results for the Semantic-Adaptive Motion Response module on the INSTA dataset.}
	{
		\begin{tabular}{lllll}
			\toprule
			Method             & PSNR\(\uparrow\) & SSIM\(\uparrow\) & LPIPS\(\downarrow\) & \(L_1^*\)\(\downarrow\)\\ \midrule
			w/o geometry residual   & 28.31 & 0.9461 & 0.0487 & 1.011 \\
            w/o semantic routing  & 30.21 & 0.9551 & 0.0366 & 0.877 \\
			w/o uv displacement   & 30.19 & 0.9542 & 0.0369 & 0.870 \\
			w/o semantic-adaptive mod.  & 30.32 & 0.9552 & 0.0357 & 0.876 \\
			\midrule
			Ours               & \cellcolor{c1}30.50 & \cellcolor{c1}0.9579 & \cellcolor{c1}0.0307 & \cellcolor{c1}0.806 \\
			\bottomrule
		\end{tabular}
	}
	\label{tab:ablation}
\end{table}

\begin{figure}[t]
	\centering
	\includegraphics[width=1.0\linewidth]{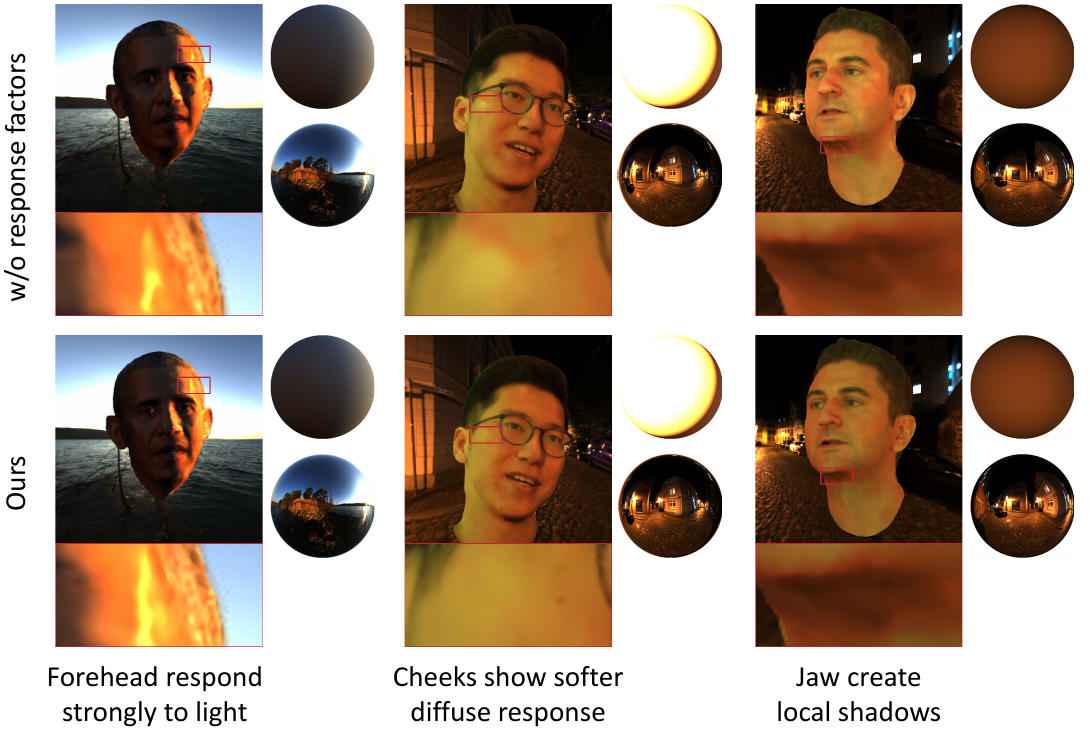}
	\caption{Ablation of the semantic response factors. Removing the response factors still allows the PBR renderer to produce plausible relighting, while the full model yields more faithful local shading and highlight variations under novel illumination.}
	\label{fig:ablation_factor}
\end{figure}

The ablation results show that each motion-response component contributes to local expression fidelity, with the geometry residual playing the most critical role. As reported in Tab.~\ref{tab:ablation}, removing the geometry residual causes the largest performance drop. This degradation is also evident in Fig.~\ref{fig:ablation}, where forehead wrinkles become less pronounced and the eyelids exhibit noticeable artifacts, as the avatar must rely solely on coarse mesh-bound motion without local Gaussian residuals. Removing either the UV displacement or semantic routing leads to milder but consistent metric degradation. Qualitatively, these variants produce smoother forehead details and less sharp tooth boundaries, indicating that frame-dependent UV motion cues and semantic feature routing are both important for modeling semantic-specific deformations. Preserving the semantic masks while removing semantic-adaptive modulation maintains spatial partitioning; however, the shared modulator still limits semantic-specific transformations, resulting in a smaller yet consistent performance drop.

To further assess expression fidelity and detail recovery for each facial semantic class, we parse each ground-truth face image using BiSeNet~\cite{Yu-ECCV-BiSeNet-2018} and compute PSNR and SSIM over the pixels associated with each class. Under this protocol, Fig.~\ref{fig:ablation_region} further validates the importance of semantic-adaptive motion response modeling: the full model achieves the highest PSNR and SSIM across the evaluated eye, nose, mouth, neck, and face classes, with particularly clear gains in expressive areas such as the mouth and eyes. This trend is consistent with Fig.~\ref{fig:ablation}: semantic routing and semantic-adaptive modulation help preserve local structures, while the geometry residual and UV displacement provide the deformation capacity and frame-dependent guidance required to align these structures under expression changes.

\subsubsection{Semantic-Adaptive Illumination Response}
We then ablate the key components of the Semantic-Adaptive Illumination Response module. Since the semantic response factors directly control how different facial semantics modulate diffuse and specular reflectance under illumination, we evaluate them separately from the stabilization terms used for intrinsic decomposition. This separation allows us to examine two complementary aspects: how the response factors influence semantic-dependent relighting behavior, and whether the remaining training design stabilizes the estimation of materials, normals, and illumination under monocular supervision.

Fig.~\ref{fig:ablation_factor} shows that the semantic response factors provide a modest but consistent improvement in relighting realism. Without these factors, the renderer can still capture the dominant illumination effects through the estimated normals, materials, and environment lighting. However, the enlarged areas reveal less well-calibrated local responses: forehead highlights become weaker and less complete under strong directional light, cheeks exhibit less smooth diffuse shading, and the jaw area shows less coherent local shadow variations. With the response factors, these effects better match the material-dependent semantic behavior illustrated in Fig.~\ref{fig:motivation}. Therefore, the response factors do not introduce an additional lighting mechanism; rather, they mildly calibrate the diffuse and specular responses of different facial semantics, allowing the existing PBR shading model to better capture semantic-dependent appearance variations.

\begin{figure}[t]
	\centering
	\includegraphics[width=1.0\linewidth]{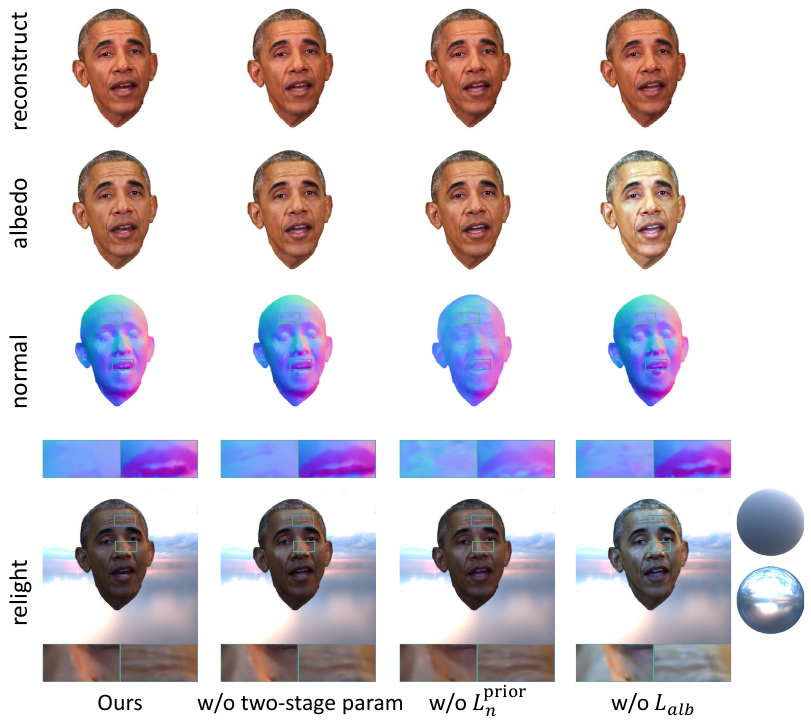}
	\caption{Ablation of intrinsic decomposition and regularization components.}
	\label{fig:ablation_relight}
\end{figure}

As shown in Fig.~\ref{fig:ablation_relight}, we further evaluate several components designed to stabilize intrinsic decomposition. Without the two-stage parameterization, canonical normals and response factors are optimized prematurely, allowing normals and illumination responses to compensate for early appearance errors and thereby producing local artifacts. Removing the normal prior causes the most severe degradation in normal estimation, leading to overly smooth facial contours and weaker structured illumination changes under novel lighting. Removing the albedo loss produces over-bright albedo and encourages baked-in illumination, which in turn makes the relit face less consistent with the target environment.

\begin{figure}[t]
	\centering
	\includegraphics[width=1.0\linewidth]{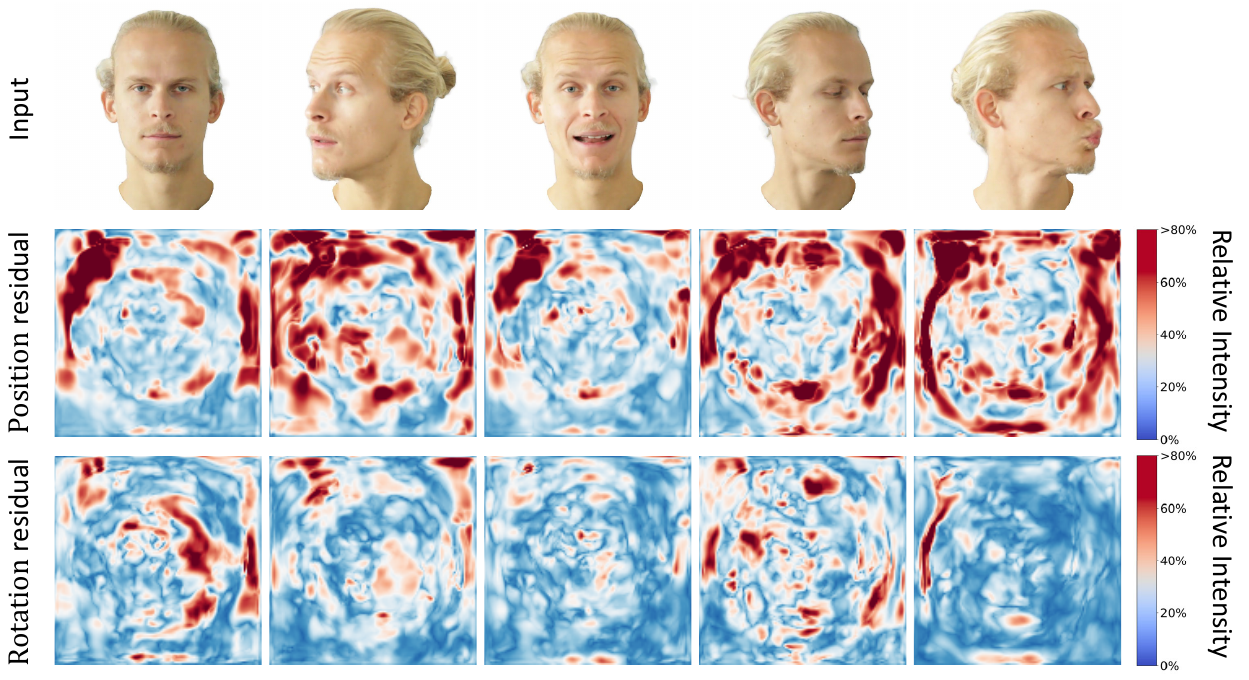}
	\caption{Visualization of UV position and rotation residuals predicted by SAMIRA under different expressions and poses. Warmer colors indicate stronger relative residual magnitudes.}
	\label{fig:residual}
\end{figure}

\subsection{Component Analysis}
We visualize the UV residual maps predicted by the Semantic-Adaptive Motion Response module in Fig.~\ref{fig:residual}. The residuals remain nonzero even under a neutral expression, particularly in hair-related areas whose geometry is difficult to represent using the coarse FLAME mesh alone. As the head pose changes, the position residuals become stronger and more spatially extensive, which is expected since pose variation introduces larger spatial discrepancies between the mesh prior and the observed appearance. More importantly, expression changes activate residuals in semantically relevant areas: mouth-related expressions produce stronger responses around the lips and jaw, whereas gaze and eyelid motions induce localized residual variations near the eyes. These patterns suggest that the network does not predict arbitrary offsets; instead, it learns residual corrections that are correlated with both pose-induced misalignment and semantic-specific facial motion.
\section{Conclusion}
We presented SAMIRA, a 3D Gaussian avatar framework for semantic-adaptive motion-illumination response modeling from monocular videos. The key idea is to model motion-illumination responses as semantic-specific avatar behaviors rather than globally shared responses: motion response captures semantic-specific deformation beyond coarse mesh binding, while illumination response captures semantic-dependent reflectance changes under novel lighting. The proposed \textit{Semantic-Adaptive Motion Response} module predicts localized Gaussian residuals in a topology-consistent UV space, and the \textit{Semantic-Adaptive Illumination Response} module learns compact diffuse and specular response factors integrated into deferred PBR Gaussian rendering. Experiments on self-reenactment, cross-reenactment, and relighting tasks demonstrate that SAMIRA improves expression fidelity and relighting realism.

\bibliographystyle{IEEEtran}
\bibliography{main.bib}

\raggedbottom

\end{document}

\title{Supplementary Material for\
Relightable 3D Avatar Reconstruction with Semantic-Adaptive Motion-Illumination Responses}

\markboth{Journal of \LaTeX\ Class Files,~Vol.~18, No.~9, September~2020}%
{How to Use the IEEEtran \LaTeX \ Templates}

\maketitle

\section{Overview}
This supplementary material provides additional experimental results and discussions for SAMIRA. In Sec.~\ref{sec:general}, we compare SAMIRA with representative generalizable avatar reconstruction methods to further clarify the trade-off between feed-forward reconstruction and subject-specific optimization. In Sec.~\ref{sec:limitations}, we discuss the current limitations of SAMIRA, including its dependence on monocular supervision, template-guided tracking, and compact illumination response modeling. Finally, Sec.~\ref{sec:ethics} presents the ethical considerations related to controllable avatar reconstruction and relighting.

\section{Additional Experimental Results}

\subsection{Comparison with Generalizable Methods}
\label{sec:general}
We further compare SAMIRA with representative generalizable avatar reconstruction methods, including LAM~\cite{he2025lam} and GAGAvatar~\cite{chu2024gagavatar}. These methods can reconstruct plausible avatars from very limited input and are more efficient for unseen identities. In contrast, SAMIRA requires subject-specific optimization from a monocular video, aiming for higher reconstruction fidelity and more accurate expression modeling. This comparison is therefore intended to analyze the trade-off between generalization efficiency and per-subject reconstruction quality.

As shown in Tab.~\ref{tab:general}, SAMIRA achieves higher PSNR and SSIM, as well as lower LPIPS and \(L_1^*\), on the self-reenactment task. These results suggest that, given sufficient monocular observations, per-subject optimization can better fit identity-specific appearance and expression dynamics than category-level priors alone.

\begin{table}[th]
	\centering
	\caption{Quantitative self-reenactment comparison with representative generalizable avatar reconstruction methods.}
	{
		\begin{tabular}{lllll}
			\toprule
			Method       & PSNR\(\uparrow\) & SSIM\(\uparrow\) & LPIPS\(\downarrow\) & \(L_1^*\)\(\downarrow\)\\ \midrule
			LAM          & 21.71 & 0.8655 & 0.1292 & 2.744 \\
			GAGAvatar    & 22.81 & 0.8942 & 0.1042 & 2.198 \\
			\midrule
			Ours         & \cellcolor{c1}30.50 & \cellcolor{c1}0.9579 & \cellcolor{c1}0.0307 & \cellcolor{c1}0.806 \\
			\bottomrule
		\end{tabular}
	}
	\label{tab:general}
\end{table}

\begin{figure}[th]
\centering
\includegraphics[width=1\linewidth]{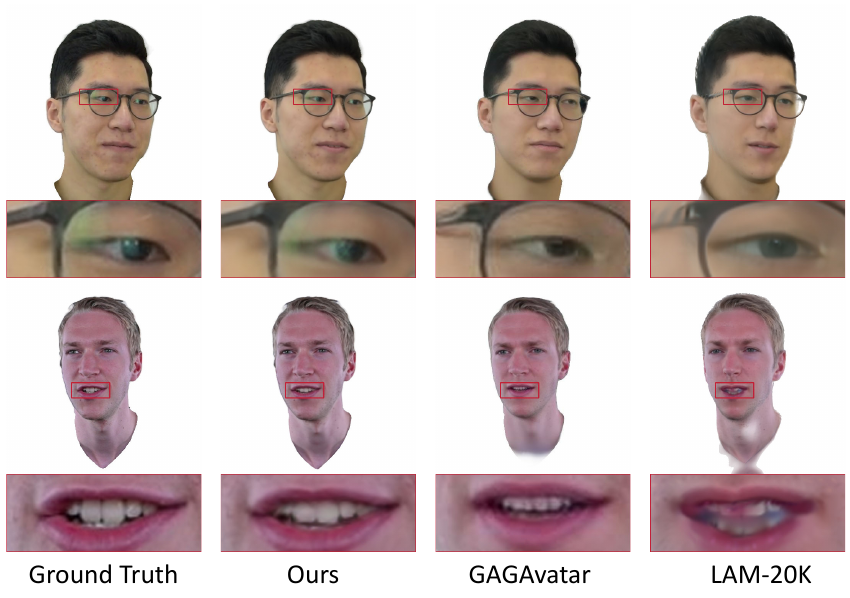}
\caption{Qualitative self-reenactment comparison with generalizable methods. Generalizable methods provide plausible feed-forward reconstruction, while subject-specific optimization better preserves identity details and local expression motion.}
\label{fig}
\end{figure}

The qualitative comparisons in Fig.~\ref{fig} show a consistent trend. Generalizable methods produce reasonable overall geometry, but their results are often smoother in fine-grained regions. By optimizing on the target monocular video, SAMIRA better preserves identity-specific structures, including facial contours, wrinkles, hair boundaries, and accessories. These observations further indicate that SAMIRA complements generalizable avatar reconstruction methods by focusing on high-fidelity subject-specific modeling rather than feed-forward efficiency.

\section{Limitations and Discussion}
\label{sec:limitations}
SAMIRA improves semantic-adaptive motion-illumination response modeling from monocular videos, but it still inherits limitations from its monocular and template-guided setting. The motion response module depends on FLAME tracking, a fixed UV layout, and predefined semantic masks. Therefore, inaccurate tracking or subtle motions around the lips and eyelids may still lead to residual misalignment. The illumination response module approximates semantic-dependent reflectance using compact diffuse and specular response factors, which improves stability but does not explicitly model cast shadows, interreflection, or extreme illumination changes. More broadly, monocular RGB supervision leaves geometry, material, normals, and lighting under-constrained.

These limitations suggest several possible directions for future work. More robust tracking and adaptive semantic partitioning may improve local motion modeling in highly deformable regions. Incorporating stronger physical constraints, temporal illumination consistency, or lightweight multi-view cues could further reduce the ambiguity among geometry, material, and lighting. In addition, more expressive illumination response models may help handle complex lighting effects while preserving the efficiency and stability of the current compact formulation.

\section{Ethics Statement}
\label{sec:ethics}
All datasets used in this study are publicly available and used in accordance with their licenses. SAMIRA supports benign applications such as telepresence, digital content creation, virtual communication, and controllable avatar editing. However, it could also be misused to generate controllable synthetic portraits or videos without consent. We explicitly oppose malicious or unauthorized use, including identity fraud, impersonation, or deceptive media creation. Responsible deployment should require informed consent, clear disclosure of synthetic content, and compliance with relevant legal and ethical guidelines.

\bibliographystyle{IEEEtran}
\bibliography{main.bib}